# Tumor Saliency Estimation for Breast Ultrasound Images via Breast Anatomy Modeling

Fei Xu[1], Yingtao Zhang[2], Min Xian[3], H. D. Cheng[1,2], Boyu Zhang[1],

Jianrui Ding[4], Chunping Ning[5], Ying Wang[6]

[1] Department of Computer Science, Utah State University, Logan, USA

[2] School of Computer Science and Technology, Harbin Institute of Technology, Harbin, China

[3] Department of Computer Science, University of Idaho, Idaho Falls, USA

[4] School of Computer Science and Technology, Harbin Institute of Technology, Weihai, China

[5] Department of Ultrasound, The Affiliated Hospital of Qingdao University, Qingdao, China

[6] Department of General Surgery, The Second Hospital of Hebei Medical University, Shijiazhuang, China

**Abstract**

Tumor saliency estimation aims to localize tumors by modeling the visual stimuli in medical images. However, it is a challenging task for breast ultrasound due to the complicated anatomic structure of the breast and poor image quality; and existing saliency estimation approaches only model generic visual stimuli, e.g., local and global contrast, location, and feature correlation, and achieve poor performance for tumor saliency estimation. In this paper, we propose a novel optimization model to estimate tumor saliency by utilizing breast anatomy. First, we model breast anatomy and decompose breast ultrasound image into layers using Neutro-Connectedness; then utilize the layers to generate the foreground and background maps; and finally propose a novel objective function to estimate the tumor saliency by integrating the foreground map, background map, adaptive center bias, and region-based correlation cues. The extensive experiments demonstrate that the proposed approach obtains more accurate foreground and background maps with the assistance of the breast anatomy; especially, for the images having large or small tumors; meanwhile, the new objective function can handle the images without tumors. The newly proposed method achieves state-of-the-art performance when compared to eight tumor saliency estimation approaches using two breast ultrasound datasets.

## 1. Introduction

Breast cancer is the most commonly diagnosed cancer. More than 2 million women every year are diagnosed with breast cancer, and more than 620,000 will die from the disease [1]. Early detection and treatment of breast cancer can substantially increase the survival rate of diagnosed women [2, 3]. In fact, the five-year relative survival rate is almost 100% if cancers are diagnosed and treated at stages 0 and 1; however, the rate is only 22% for women with advanced-stage breast cancers, highlighting the critical need for the methodologies of early detection.

In clinical routine, breast ultrasound (BUS) is a primary modality for cancer screening [4, 5], and automatic BUS image segmentation methods are essential for cancer diagnosis and treatment planning. In the last decade, many automatic BUS segmentation approaches have been studied [5-11]. The primary strategy of the approaches is to locate tumors automatically by modeling domain-related priors. Some strong constraints such as the number of tumors, tumor size, and predefined tumor locations, were utilized and resulted in dramatic performance degradation when BUS images were collected under different settings or situations such as low contrast, more artifacts, containing no tumor/more than one tumors per image, etc. Therefore, it is crucial to develop automatic BUS segmentation techniques that are invariant and robust to such settings or situations.

Visual saliency estimation (VSE) measures the degrees of human's attention attracted by different image regions and is essential and accessible to achieve automatic image segmentation [7, 12-37]. In cognitive science and computer vision, visual saliency is a property that makes objects stand out visually from the neighbors. It is a robust feature, and modeling VSE can help to detect target objects automatically and accurately. In BUS images, tumors typically attract the attention of radiologists even under very different imaging conditions. Examples of applying VSE to BUS images are shown in Fig. 1. Many approaches [7, 26, 34] were proposed to model the visual cues attracting radiologists' attention. In [7], Shao *et al.* proposed a model based on saliency estimation for fully automatic tumor detection. The model combined tumor prior knowledge and human visual saliency estimation hypothesis and achieved very good performance using their own BUS image dataset. However, it had two main drawbacks: 1) always outputs a salient region even there was no tumor in the image (Fig. 2 (b)); and 2) could not deal with the images having large tumors, shadows, and low contrast (Fig. 1 (c)). Xie *et al*. [26] computed tumor saliency by comprising intensity, blackness ratio, and superpixel contrast separately; and the average of the values of the three components was the final saliency value of each pixel. The drawbacks were shared as [7] due to the nature of direct mapping and the strategy of "winner-take-all". Xu *et al*. [27] proposed a general bottom-up saliency estimation model that integrated the robust hypotheses: the global contrast, adaptive center-bias, boundary constraint and the smoothness term based on color statistic. The model was flexible, and the global optimum could be reached by using the primal-dual interior point method. However, the model could not deal with low contrast or gray-level images; furthermore, it always located a salient region and could not handle images without salient objects. Recently, Xu *et al*. [34] proposed a novel hybrid framework for tumor saliency estimation. In the framework, it integrated the background map, foreground map [5] and adaptive center-bias. However, it shared the same drawback as [27] that the data term in the objective function only penalized pixels with nonzero saliency values; and the equality constraint forced the summation of all saliency values to be 1 that led to at least one relative salient object in every image (Fig. 2 (c)).

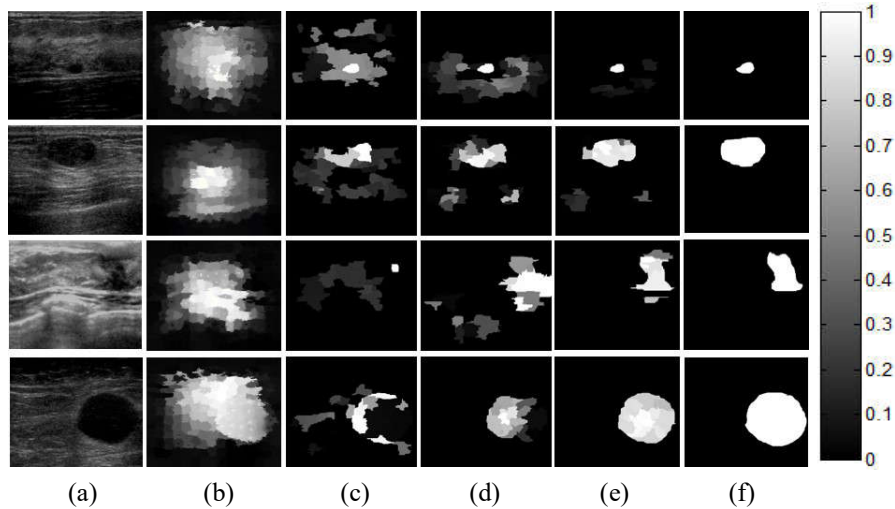

Fig. 1. Visual saliency estimation for BUS images. (a) Four original BUS images; (b-d) results of [25], [7], and [34], respectively; (e) results of the proposed method; and (f) the ground truth (GT). The region with higher intensity indicates the region has higher possibility belonging to a tumor.

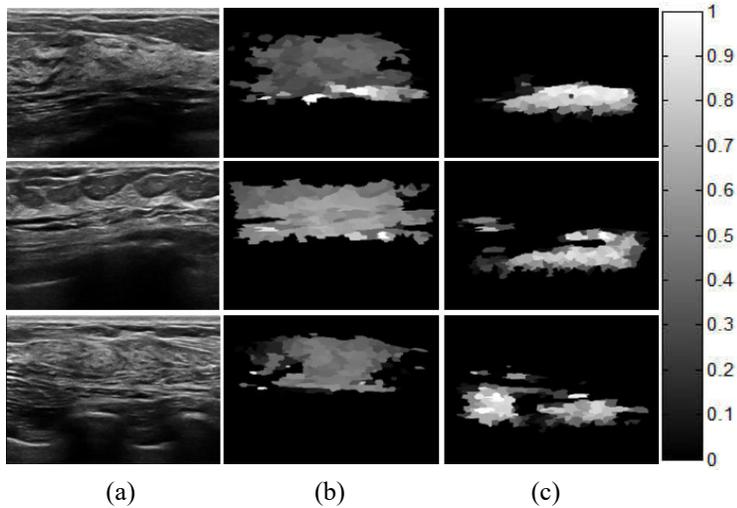

Fig. 2. The methods [7] and [34] always generate salient regions in the images without tumors. (a) Original BUS images without tumors, (b) and (c) the saliency maps generated by using method [7] and [34], respectively. The region with higher intensity indicates the region with higher possibility belongs to a tumor.

To overcome the above challenges, we propose a novel optimization-based approach for estimating tumor saliency map of BUS image. First, we construct a new cost function that penalizes the inconsistency between image features and saliency values for both salient and non-salient pixels. By doing so, the equality constraint [27, 34] can be eliminated, and the new approach does not output salient regions for every BUS image. Second, breast anatomy is modeled by using Neutro-Connectedness theory [38, 39] and applied as non-local context information to solve the problem of outputting wrong salient regions for BUS images with dark shadows (see Fig. 8). The tumor regions will have higher connectednesses than that of the background in the low contrast images. The results will be much more reliable by utilizing the breast anatomy knowledge, and it makes the shadows layer with high rate be background; especially, for the images having large tumors. The framework of the proposed method is shown in Fig. 3.

The rest of the paper is organized as follows: in section 2, the details of the problem formulation and components of the new cost function are discussed; section 3 is the implementation of optimizing the proposed approach; section 4 explains experimental result; and section 5 discusses the conclusion and future work.

## 2. Proposed method

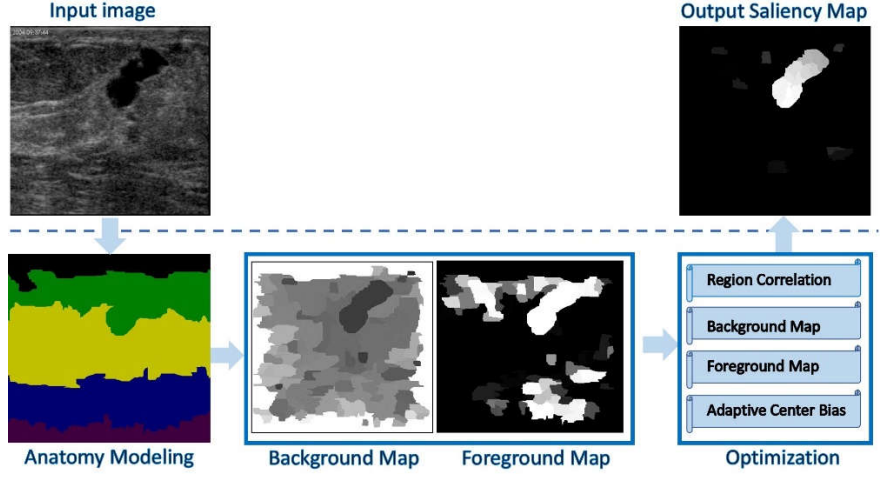

Fig. 3. Pipeline of the proposed approach.

**2.1. Problem formulation**

In the proposed approach, tumor saliency estimation (TSE) is formulated as a Quadratic Programming (QP) problem, and we focus on solving the problems in existing approaches by building a united optimization-based framework that incorporates robust cognitive hypothesis, e.g., the adaptive center-bias, and region-based correlation hypothesis, and the background and foreground cues.

Let $\{R_i\}_{i=1}^{i=N}$ be a set of image regions generated by a quick shift algorithm [40], and $S = (s_1, s_2, \cdots, s_N)^T$ be a vector of saliency values, where $s_i$ denotes the saliency value of the $i$th region and $s_i \in [0, 1]$. The TSE problem is formulated as

$$\begin{aligned}\text{minimize } & E(S) = \alpha E^{data}(S) + E^{smooth}(S) \\ \text{subject to } & 0 \le s_i \le 1, i = 1, 2, \cdots, N; \\ & B^T S = 0, B = (b_1, b_2, \cdots b_N)^T, b_i = \{0,1\}\end{aligned} \quad (1)$$

where the data term $E^{data}$ models the background cue, foreground cue and adaptive center-bias cue; and the smoothness term $E^{smooth}$ models the region-based correlation; $\alpha$ balances the influence of the two terms; the equality constraint $B^T S = 0$ is only applied to mask border regions; and $b_i$ is 1 if the $i$th region is adjacent to the image border, and 0 otherwise.

$$E^{data}(S) = E^{fg}(S) + E^{bkg}(S) \quad (2)$$

$$E^{fg}(S) = S^T \cdot (-(\ln(D) + \beta \ln(W))) \quad (3)$$

$$E^{bkg}(S) = (1-S)^T \cdot (-\ln(T)) \tag{4}$$

In Eq. (2), $E^{fg}$ defines the cost of assigning non-zero saliency value to each image region, and $E^{bkg}$ defines the cost of assigning zero value to an image region. In previous optimization-based approaches [34], only $E^{fg}$ was defined, and no explicit cost was given for outputting zero saliency values. In order to avoid the configuration of all zero saliency values for the entire image, a constraint $\sum s_i = 1$ was defined to force the output to have at least one salient region for every image. This is one of the major drawbacks of previous approaches and makes them unable to deal with BUS images without tumors. In order to overcome the drawback, $E^{bkg}$ is defined in the data term of the cost of assigning zero value to an image region. This strategy can avoid the zero-configuration problem, because all zeros will lead to a high penalty if a salient region (tumor) exists in the image; and it outputs all zeros only when no tumor exists. In Eqs. (3)-(4), $W = (w_1, w_2, \cdots, w_N)^T$ is the foreground map, and defines the possibility of each image region to be a tumor region; $D = (d_1, d_2, \cdots, d_N)^T$ is the distance map, and $d_i$ defines the distance between the $i$th region and the adaptive center; and $\beta$ balances the contribution of the two terms. $T$ denotes the background map, and defines the possibility of an image region to be a background region. The definitions of $W$, $D$ and $T$ will be given in section 2.2.

$$E^{smooth}(S) = \sum_{i=1}^{N} \sum_{j=1}^{N} (s_i - s_j)^2 r_{ij} Dist_{ij} \tag{5}$$

$E^{smooth}$ in Eq. (5) defines the penalty on similar regions with different saliency values. The terms $r_{ij}$ and $Dist_{ij}$ define the similarity and spatial distance between the $i$th and the $j$th regions, respectively.

The problem defined by Eqs. (1) – (5) is a typical QP problem with linear equality and inequality constraints. The original problem can be rewritten as follows:

$$\begin{aligned} \text{minimize } f_0(S) = &\alpha \sum_{i=1}^{N} -s_i \ln(d_i) + \gamma \sum_{i=1}^{N} -s_i \ln(w_i) + \\ &\alpha \sum_{i=1}^{N} -(1-s_i)\ln(t_i) + \sum_{i=1}^{N}\sum_{j=1}^{N}(s_i - s_j)^2 r_{ij} Dist_{ij} \\ \text{subject to } & 0 \leq s_i \leq 1, i = 1,2,\cdots,N; \\ & B^T S = 0, B = (b_1, b_2, \cdots b_N)^T, b_i = \{0,1\} \end{aligned} \tag{6}$$

where $\gamma = \alpha\beta$, refer Eqs. (1) and (3).

## 2.2 Data term

The data term is composed of three major components: foreground map (*W*), distance map (*D*), and background map (*T*). The foreground map models priors of general tumor appearance; the distance map models the adaptive center-bias hypothesis; and the background map is defined as the weighted connectedness between image and border regions. The definitions of the three parts are guided by breast anatomy.

### 2.2.1 Breast anatomy modeling using Neutro-Connectedness

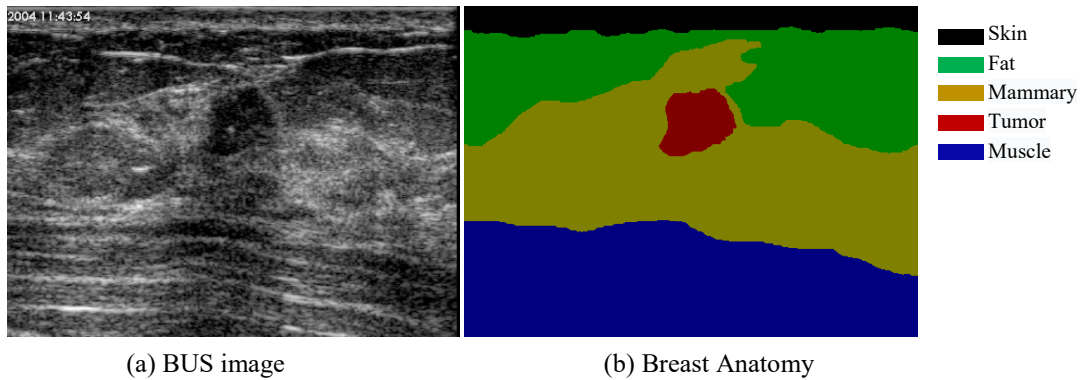

(a) BUS image  (b) Breast Anatomy

Fig. 4. An example of breast anatomy.

Breast anatomy represents the structure of the breast and is useful for breast tumor detection and classification in clinical practice. Breast contains four major layers: skin and fat layer, mammary layer, and muscle layer, and breast tumor mainly exists in mammary layer (see Fig.4). Duo to biological properties, regions in different layers have different appearances. In this work, we model breast anatomy by a new Neutro-Connectedness (NC)-based model [34] that incorporates the depth information of breast regions; and decompose BUS images into 3 to 5 layers according to NC paths.

There are two components in NC: the degree of connectedness *t* and confidence of connectedness *c*, $NC(i,j) = [t(i,j), c(i,j)]$ where *i* and *j* indicate the *i*th and *j*th pixel or region, respectively. Image regions from the same layer should have strong connectedness (e,g., high *t* and *c* values), and from different layers have weak connections. NC builds on the following three fundamental concepts:

(1) NC *of two adjacent regions i* and *j*. The degree of connectedness of two adjacent regions is defined as their similarity, noted as $\mu_t$; and the degree of confidence is defined as the homogeneity between them, noted as $\mu_c$.

(2) NC *of a path*. The degree of connectedness of a path is defined as the minimum value of $\mu_t$ along the path, and degree of confidence is the minimum $\mu_c$ value along the path.

(3) NC *of any two regions*. The degree of connectedness is defined by the strongest path between the two regions. It uses the confidence of the corresponding path as the degree of connectedness confidence of the two regions.

Breast anatomy was utilized in [5, 7] for tumor segmentation. Shao *et al.* [7] identified two horizontal lines to remove the fat and muscle by applying phase congruency [41] and Otsu's thresholding. The horizontal lines are difficult to identify accurately; in some cases, part of the tumor could be separated into the fat region. Xian *et al.* [5] detected tumors by attenuating the dark regions of fat and shadows; the approach was not sensitive to small tumors. One common drawback of the two approaches is the assumption of the necessary existence of a tumor in each BUS image.

In this work, we redefine the NC *of two adjacent regions* by utilizing the region similarity and depth. The new depth term has the additional constraint of the growth of NC along the vertical direction.

$$u_t(i,j,k) = r_{ij}(i,j) \cdot o_{ik}(i,k) \quad (7)$$

$$\mu_c(i,j) = min(h(i), h(j)) \quad (8)$$

$$r_{ij}(i,j) = exp(-|I(i) - I(j)|/\sigma_1^2) \quad (9)$$

$$o_{ik}(i,k) = exp(-|row(i) - row(k)|/\sigma_2^2) \quad (10)$$

In Eq. (7), $r_{ij}$ denotes the similarity between the *i*th and *j*th regions, and $o_{ik}$ is the normalized depth difference between the *i*th region and the root region (*k*) of a NC tree. In Eq. (8), $h(\cdot)$ defines the homogeneity of a region [38, 39]. In Eq. (9), $I(i)$ and $I(j)$ are the normalized intensities of the *i*th and *j*th regions, respectively; $row(i)$ denotes the row index of the *i*th region center. $\sigma_1^2$ and $\sigma_2^2$ control the shapes of the two exponential functions. $\sigma_1^2$ is 0.5 by experiment, and $\sigma_2^2$ is initialized as 0.2 and updated adaptively to control the number of layers between three and five. If the number of layers is greater than five, decreasing $\sigma_2^2 = \sigma_2^2 - 0.05$; and increasing $\sigma_2^2 = \sigma_2^2 + 0.05$ otherwise.

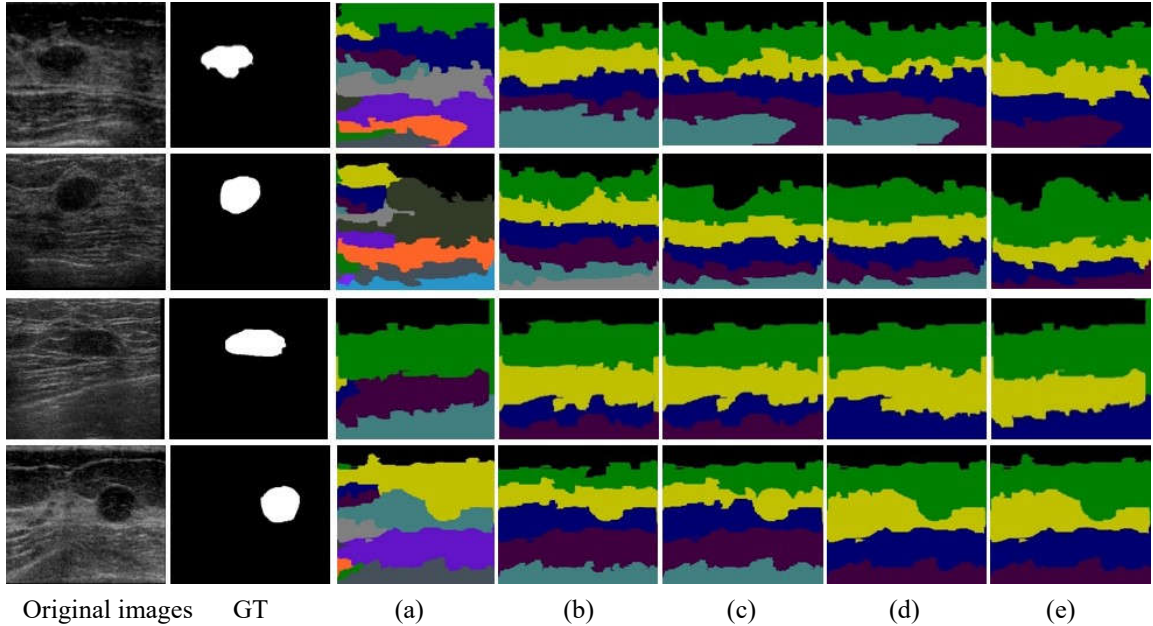

Fig. 5. Effectiveness of different $\sigma_2^2$. (a) $\sigma_2^2 = 0.2$, before merging;(b)- (e): with $\sigma_2^2 = 0.05, 0.1, 0.15$ and $0.2$, after merging, respectively. From top to bottom, the same color indicates the same layer.

After computing the NC *of two adjacent regions*, the connectednesses of a path and between any two regions can be calculated easily. The left-side boundary regions of an image are set as the roots for generating NC. All the regions on a path are in a group (layer). If a layer cannot cover more than ¾ of the image width, it will be merged into its nearest layer. The effectiveness of the merging step with different $\sigma_2^2$ is shown in Figs. 5. (a) and (e). Note that each generated image layer is composed of a group of image regions that have high connectedness with each other; those regions has high possibililty from a same biological tissue layer, but the generated image layer is not the biologic tissue layer.

**2.2.2 Foreground map (FG) generation**

Foreground map (FG) measures image regions' possibilities to be tumor regions. We propose a new method to generate the foreground map by using both image appearance and breast anatomy. The Z-shaped function is used for each layer to emphasize image regions with low intensities, and layer's location generated in section 2.2.1 is employed to reduce the impact of the dark regions from the fat and shadow regions. The Z-shaped function in [5] is utilized; however, the parameters *a*, *b*, *c* in Eq. (11) are chosen adaptively for different layers of different images, and the image intensities are mapped in [0,1].

$$Z(I(i); a, b, c) = \begin{cases} 1, & \text{if } I(i) \leq a \\ 1 - \frac{(I(i)-a)^2}{(c-a)(b-a)}, & \text{if } a < I(i) \leq b \\ \frac{(I(i)-c)^2}{(c-a)(c-b)}, & \text{if } b < I(i) \leq c \\ 0, & \text{if } I(i) > c \end{cases} \quad (11)$$

Since the fat layer is darker than other regions, if using the unified parameters *a,b* and *c* in Eq. (11) to generate the foreground map, it will cause the fat layer to have the highest intensity after Z function and to miss a part of the tumor in some cases. Therefore, we combine global *a, b* and *c* with the local ones together to generate the foreground map of each layer. The global parameters are used to decrease the intensities of high-intensity smooth layers. We define the global value *a* as $a_g = I^{1/10}$, the global *c* as $c_g = I^{6/10} - a_g$, and the mean value of the intensities less than $c_g$ as $b_g$. And *I* is the intensity list of the entire image regions and $I^{1/10}$ is the one-tenth of *I* in an ascending order. The local values of $a_i$ and $c_i$ in the *i*th layer are defined as $a_i = Glist_i^{1/10}$ and $c_i = Glist_i^{6/10}$, and $b_i$ is the mean intensity of all the intensities less than $c_i$. $Glist_i$ is the intensity list of the region intensities in the *i*th layer. We propose Algorithm 1 to generate the foreground map of the *i*th layer.

**Algorithm 1**: Generating the initial FG of the *i*th layer

---

**Input**: $I, a_g, c_g, layerM, i$; *I* is the intensity list of the regions, $layerM$ contains the breast anatomy information
**Output**: $flag$, FG, $layerM$; $flag$ indicates whether there is a dark layer
1. Calculate the local $a_i, c_i,$ and $b_i$ based on $layerM$; store the local $a_i, c_i,$ and $b_i$ in $layerM$
2. If $(b_i - a_g < 0.1(c_g - a_g))$ then
3.    Apply the local parameters $a_i, c_i,$ and $b_i$ in the Eq. (11) to generate the weight map FG for the layer
4.    $flag$ =1;
5. Else
6.    If $a_i > c_g$, which means most of the regions in the layer have high intensities, assign the intensity of the layer in the weighted map 0 and $flag$ =-1;
7.    Else
8.      $a_i = \min(a_g, a_i), c_i = \min(c_g, c_i)$
9.      $flag$ =0
10.      Apply the global parameters in Eq.(11) to generate the initial weight map FG for the layer
11. End

In **Algorithm 1**, the *layerM* contains the layers' information: the regions list, root region's row index and local parameters of each layer. The condition $b_i - a_g < 0.1(c_g - a_g)$ is used to determine whether there is a dark layer. If $b_i$ is small enough, it indicates that most of the regions in the layer have low intensities. Based on above analysis, it will produce two results under such condition: 1) if the layer locates in the bottom or top part, it has a high possibility that the layer is a non-mammary dark layer, and the local parameters will be used to generate FG; 2) otherwise, it is a large tumor in the mammary layer with high possibility and a new global $a_g'$, $c_g'$ and $b_g'$ will be used in Eq. (11), see the details in **Algorithm 2**. The condition $a_i > c_g$ is used to check if the layer is a smooth bright layer or a normal one. If $a_i$ is larger than the global $c$, it indicates that most of the regions in the layer have high intensities; therefore, it has a very low possibility to contain a tumor.

The tumor-like regions may exist in the top or bottom layer, and there may be more than one dark layer. Thus, the layers will be separated into the bottom, top and middle parts to generate the final FG, and assign the weight for each layer. The initialized weights for the three parts are given as follows:

$$layerW_i^{LT} = layerW_i^{LB} = \max\left(\left(i - \left\lceil\frac{layerNum}{2}\right\rceil\right)^2, 1\right); \tag{12}$$

$$layerW_i^{LM} = \exp\left(\frac{-\sqrt{layerNum}}{2(loopE-loopS+1)}\right); \tag{13}$$

where $layerW_i^L$ indicates the weight for the *i*th layer which located in part $L$, $L \in \{LT, LB, LM\}$ which indicates the layer locating in the top (*LT*), bottom (*LB*) and middle (*LM*) parts, respectively. *layerNum* is the number of total layers in the image; *loopE* is the number of start layer in the bottom part and initialized as *layerNum*; and *loopS* is the number of end layer in the top part and initialized as 1. The details of generating the foreground maps and weights in different parts are described in Algorithm 2. The foreground map examples are shown in Fig.6.

**Algorithm** 2: Generating the final FG and *layerW*

---
**Input**: $I$, *layerM*
**Output**: FG, *layerW*
    1. Calculate global $a_g$ and $c_g$
---

2. $loopE = layerNum$; $loopS=1$;
3. Initial the bottom part layer range as $\left[\left\lceil\frac{layerNum+1}{2}\right\rceil, layerNum\right]$, and the stop part range as $\left[1, \left\lceil\frac{layerNum+1}{2}\right\rceil\right]$
4. Apply **Algorithm 1** to each layer in the two parts and use $a_g, c_g, i$ (from 1 to *layerNum*) and *layerM* as the input to generate $flag$ and the initial FG for each layer
5. For the bottom part, if $flag$ is 1 and there is another layer with dark flag by checking the initial FG information, assign the layer weight using Eq. (12) and extend the starting layer of the bottom part by $loopE = loopE$ -1; otherwise, stop extending the bottom part. If $flag$ is 1 and all the other layers have high local $a_i$, stop extending the bottom part and keep $loopE = layerNum$; The FG is generated using the global parameters in Eq. (11).
6. For the top part, if $flag$ is 1 and there is another layer with dark flag by checking the initial FG information, assign the layer weight using Eq. (12) and the top part ending layer as $loopS = loopS$ +1; otherwise, stop extending the top part.
7. The range of the bottom part layers is revised by (*loopE*, *layerNum*] and the range of top part layers is [*1*, *loopS*). The final FG of the bottom and top parts are the same as the initial FG.
8. To reduce the shadow layer's influence, re-assign the intensity of the shadow layer's regions to be 255 as the new $I'$, Calculate the new $a_g'$, $c_g'$. If this is a dark layer, assign $a_i' = \max(a_g', a_i)$, $c_i' = \max(c_g', c_i)$, $b_i' = (a_i' + c_i')/2$; and for the normal layer, assign $a_i' = \min(a_g', a_i)$, $c_i' = \min(c_g', c_i)$, $b_i'$ is the mean value of the intensity less than $c_i'$. Apply the new local parameters into Eq. (11) to generate the FG in the layer and assign the layer weight using Eq. (13).
9. The final FG is the dot production of FG and *layerW*.

### 2.2.3 Distance map generation

Traditional saliency estimation models usually use the image center as an important visual cue to estimate the saliency map. However, it will fail when objects are far away from the center. The approach in [27] solved this problem on natural images by estimating the adaptive center (AC) using weighted local contrast map; but the local contrast map was sensitive to noise and could not achieve good performance on BUS images. In this section, we define the AC as the weight center of the foreground map.

$$AC = \frac{\sum_{x,y}(x,y)W(x,y)}{\sum_{x,y}(x,y)}$$
$$x = 1,2,\cdots M, \text{and } y = 1,2,\cdots, N \tag{14}$$

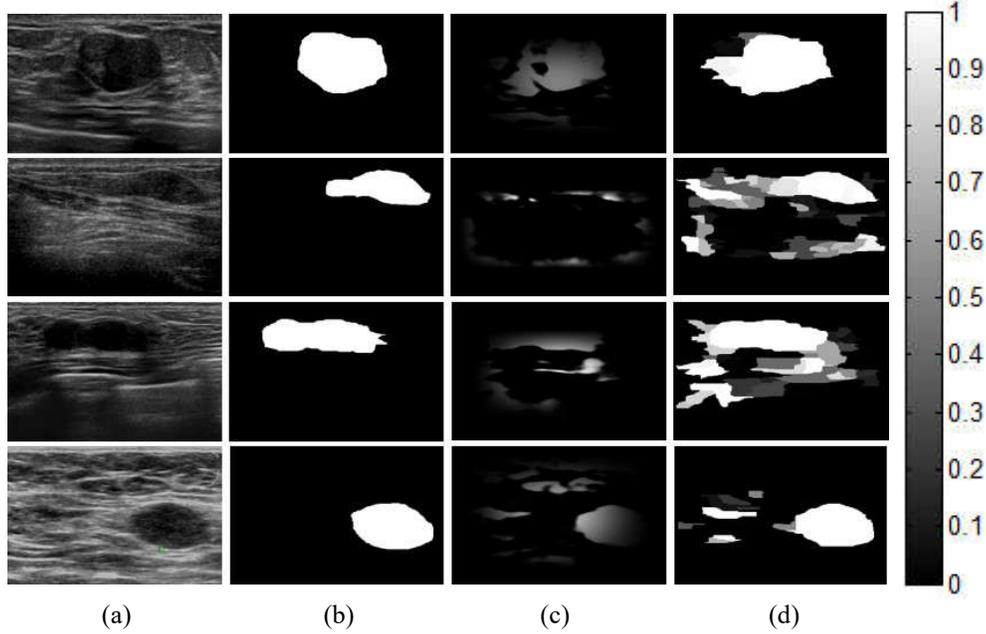

Fig. 6. Examples of FG generation. (a) Origianl BUS images; (b) the ground truth; (c) the foreground maps by [5]; and (d) the final foreground maps of the proposed approach. The region with high intensity belongs to the tumor with the higher possibility, vice versa

where $W(x,y)$ is the value of pixel $(x, y)$ in the foreground map, and M and N are the number of image rows and columns, respectively.

The AC distance vector will force the regions far away from the AC to gain small saliency value and is defined as $D = (d_1, d_2, \cdots, d_N)^T$

$$d_i = exp(-\|(x,y)_i - AC\|_2/\sigma_3^2) \qquad (15)$$

where $(x,y)_i$ is the normalized coordinates of the $i$th region's center. $\|\cdot\|_2$ is the $l_2$ norm. And $\sigma_3^2$ is set to 0.1 by experiment.

### 2.2.4 Background map (BG) generation

Boundary connectivity is an effective prior utilized in many visual saliency estimation models [27,29-33]. Most models define the boundary connectivity by using the shortest path between the local regions and the boundary. However, such connectivity could not handle noisy data well. The degree of confidence domain in NC is very useful for avoiding the fake connectedness caused by uncertainty, such as noise. As the particular characteristic that no tumor is touching the border, it sets the border regions as the background seeds to generate the NC map using the algorithm in [38], noted $nc_i$ as the NC value of the $i$th region in the NC map. The higher $nc_i$ indicates the higher possibility that the region belongs to background.

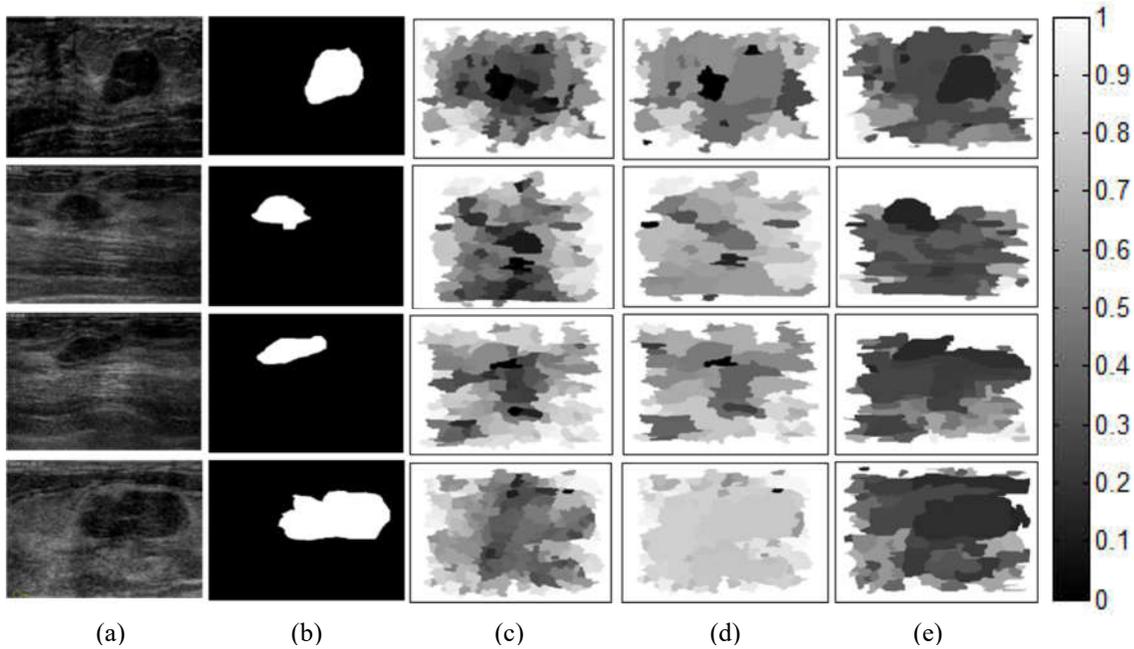

Fig.7. Examples of BG generation. (a) Original BUS images; (b) the ground truth; (c) obtainted by grapth shortest path[44]; (d) obtained by [34] without breast anatomy; and (e) obtained by the proposed method anatomy. The region with higher intensity belongs to the background with higher possibility, vice versa.

We define the value $T$ in the BG map as follows:

$$t_i = nc_i^2 \times layerW_j \tag{16}$$

where the $i$th region belongs to the $j$th layer, and $layerW_j$ is the $j$th layer's weight.

Fig. 7 shows some comparable samples. The connectedness based on graph shortest path failed to handle the BUS images with too small or too large tumors (see the 2nd-4th rows of Fig. 7 (c)), or poor quality with noise (see the 1st row). The results generated by NC without layers' information will make the tumor regions have higher connectedness than the background regions in the low contrast images (see the 2nd and 4th rows of Fig. 7 (d)). Moreover, the maps generated by the NC method are much smoother than that of graph shortest path method. The BG result will be much more reliable by utilizing NC with the layers' information, and it makes the shadow layer with a high rate to be background; especially, for the images having large tumors.

### 2.3. Smoothness term

We utilize regions' feature correlation to force similar regions to have similar saliency values. Specifically,

$$r_{ij} \cdot Dist_{ij} = r_{ij} \cdot exp\left(-\|(x,y)_i - (x,y)_j\|_2 / \sigma_1^2\right) \quad (17)$$

where $r_{ij}$ measures the similarity of regions $i$ and $j$; and $Dist_{ij}$ is defined based on the spatial distance between the $i$th and the $j$th regions; and $\|\cdot\|_2$ is the $l_2$ norm.

## 3. Optimization

The primal-dual method is applied to optimize the proposed QP problem, and the global optimal can be achieved. There are three steps to generate the optimization solution: (1) modify the Karush-Kuhn-Tucker (KKT) conditions and obtain the dual, prime and centrality residuals; (2) obtain the primal-dual search direction; and (3) update $S$ and the dual variables. The details of the optimization are described as follows:

The inequality constraints can be rewritten as a set of functions:

$$\begin{aligned} f_k(S) &= -S_k \leq 0, k = 1,2,\cdots,N \\ f_k(S) &= S_{k-N} - 1 \leq 0, k = N+1, N+2,\cdots,2N \end{aligned} \quad (18)$$

where $N$ is the number of image regions, and $S_k$ is the saliency value of the $k$th region. We write all inequality constraints in a matrix:

$$f(S) = \begin{bmatrix} f_1(S) \\ f_2(S) \\ \vdots \\ f_{2N}(S) \end{bmatrix} = \begin{bmatrix} -S \\ S-1 \end{bmatrix}_{2N \times 1} \quad (19)$$

The derivative matrix is

$$Df(S) = \begin{bmatrix} \nabla f_1(S)^T \\ \nabla f_2(S)^T \\ \vdots \\ \nabla f_{2N}(S)^T \end{bmatrix} = \begin{bmatrix} -E \\ E \end{bmatrix}_{2N \times N} \quad (20)$$

where E is the identity matrix.

The dual residual is

$$\begin{aligned}
r_d &= \nabla f_0(S)^T + Df(S)^T\lambda + \nu B \\
&= -\alpha \ln(C) - \gamma \ln(W) + \alpha \ln(T) \\
&\quad + \sum_{i=1}^{N}\sum_{j=1}^{N} 4 \times (s_i - s_j) r_{ij} \, Dist_{ij} + \begin{bmatrix} -E \\ E \end{bmatrix}^T \lambda + \nu B
\end{aligned} \tag{21}$$

where $\alpha$ and $\gamma$ balance the three terms defined in Eqs. (1) and (3); and vectors $\lambda = (\lambda_1, \lambda_2, \cdots, \lambda_{2N})^T$ and $\nu$ are the dual feasible parameters.

The primal residual is

$$r_p = \begin{bmatrix} O^T S - 1 \\ B^T S \end{bmatrix} \tag{22}$$

where $O$ is a 2N-by-1 vector, and all the values are 1s.

The centrality residual is

$$r_c = -diag(\lambda) f(S) - (1/g) O \tag{23}$$

where $g$ is the step size, and initialized as 1.

The partial derivatives of $r_d$, $r_p$ and $r_c$ with respect to variables $S$, $\lambda$ and $\nu$ are as follows:

$$\frac{\partial r_d}{\partial S} = \begin{cases} \sum_{i=1}^{N} 4 \times \left( \sum_{j=1}^{N} (r_{ij} \times Dist_{ij}) \right) - r_{ii} \times Dist_{ii}, \text{if } i = j \\ \sum_{i=1}^{N}\sum_{j=1}^{N} 4 \times (s_i - s_j) r_{ij} \, Dist_{ij}, \text{if } i \neq j \end{cases} \tag{24}$$

$$\frac{\partial r_c}{\partial S} = -diag(\lambda) \times \begin{bmatrix} -E \\ E \end{bmatrix}, \frac{\partial r_p}{\partial S} = B^T \tag{25}$$

$$\frac{\partial r_d}{\partial \lambda} = \begin{bmatrix} -E \\ E \end{bmatrix}^T, \frac{\partial r_c}{\partial \lambda} = -diag(f(s)), \frac{\partial r_p}{\partial \lambda} = 0_{1 \times 2N} \tag{26}$$

$$\frac{\partial r_d}{\partial \nu} = B, \frac{\partial r_c}{\partial \nu} = 0_{2N \times 1}, \frac{\partial r_p}{\partial \nu} = 0 \tag{27}$$

In each iteration, the Newton step $(\Delta S, \Delta \lambda, \Delta \nu)$ is obtained by solving Eq. (28) using the partial derivatives in Eqs. (24) - (27).

$$\begin{bmatrix} \frac{\partial r_d}{\partial S} & \frac{\partial r_d}{\partial \lambda} & \frac{\partial r_d}{\partial \nu} \\ \frac{\partial r_c}{\partial S} & \frac{\partial r_c}{\partial \lambda} & \frac{\partial r_c}{\partial \nu} \\ \frac{\partial r_d}{\partial \nu} & \frac{\partial r_p}{\partial \lambda} & \frac{\partial r_p}{\partial \nu} \end{bmatrix} \begin{bmatrix} \Delta S \\ \Delta \lambda \\ \Delta \nu \end{bmatrix} = -\begin{bmatrix} r_d \\ r_c \\ r_p \end{bmatrix} \tag{28}$$

The variables $S$, $\lambda$ and $\nu$ are updated using the following equations.

$$S^{k+1} = S^k + g^k \times \Delta S, \lambda' = \lambda + g^k \times \Delta\lambda, \quad (29)$$
$$v' = v + g^k \times \Delta v$$

In Eq. (29), $g^k$ is the step size and updated by using the line search method in each iteration; $g^0$ and $S^0$ are initialized as 1 and $(1/N)(1,1,\cdots,1)^T$, respectively. The dual residual, primal residual, and centrality residual are updated in each iteration, and the optimization stops when the sum of the $l_2$ norms is less than $10^{-6}$.

## 4. Experimental results

### 4.1 Datasets, metrics and setting

We validate the performance of the newly proposed method using a dataset containing 562 BUS images from a public benchmark [43] and a private dataset of 96 BUS images without tumors [34]. All experiments are conducted by using Matlab (R2018a, MathWorks Inc., MA) on a Windows-based PC equipped with a dual-core (3.6 GHz) processor and 8 GB memory.

**Metrics of saliency estimation**: Precision-recall (P-R) curve, mean Precision and Recall rate, $F_{meansure}$ and mean absolute error (*MAE*) are employed. For each method, it normalizes the intensities of the saliency map into [0, 255]. After binarization of the saliency map with thresholds ranging from 0 to 255; and it computes the precision and recall rates by comparing the thresholding result with the ground truth, and the P-R curve is calculated by averaging precision-recall ratios of the datasets. The precision and recall ratios are defined as follows:

$$Precison = \frac{|SM \cap GT|}{|SM|}, Recall = \frac{|SM \cap GT|}{|GT|}$$

where *SM* denotes the binary saliency map, *GT* is the ground truth, and $|\cdot|$ denotes the number of pixels of 1s. To obtain the average precision and recall ratios, it uses an adaptive thresholding method [16], which chooses two times the mean saliency value as the threshold. The $F_{meansure}$ [17] and *MAE* [29] are defined as

$$F_{meansure} = \frac{(1+\theta^2) Precison \cdot Recall}{\theta^2 \cdot Precison + Recall}$$

$$MAE = \sum_{i=1}^{P}|S(p_i) - G(p_i)|$$

where $\theta^2$ is set to 0.3 suggested in [17], $p_i$ is the coordinate of the $i$th image pixel, $S(p_i)$ is the saliency value of the $i$th pixel, and $G$ is the binary ground truth. The value of each pixel in $S$ or $G$ is between 0 to 1. A better algorithm will obtain a smaller $MAE$ and a larger $F_{meansure}$.

**Parameter setting:** all the experiments are based on the parameters: $\alpha = 4, \gamma = 40$.

### 4.2. Parameters tuning

**Values of $\alpha$ and $\gamma$.** As presented in section 2.1, the detection framework has 4 major parts. Applying one of the data terms cannot always provide the correct information to generate the saliency map (see Figs. 6-7.). The tuning parameter $\alpha$ controls the relative impact of the data term and smoothness term on the optimization. And $\gamma$ controls the balance effect of foreground cue and background cue. It evaluates the performance of the proposed method with $\alpha$ ranging from 0 to 100 and $\gamma$ ranging from 0 to 3000, using randomly selected subset of 60 images. There are three stages to choose the parameters. In the first stage, it makes the step size of $\alpha$ and $\gamma$ be 50 and 500 respectively and obtains the range of each parameter which can achieve better P-R curve performance and $MAE$ value if the P-R curve is similar. In the second stage, the $\alpha$ step size is 20, and the $\gamma$ step size is 100. And in the third stage, the $\alpha$ step size is 2, and the $\gamma$ step size is 20.

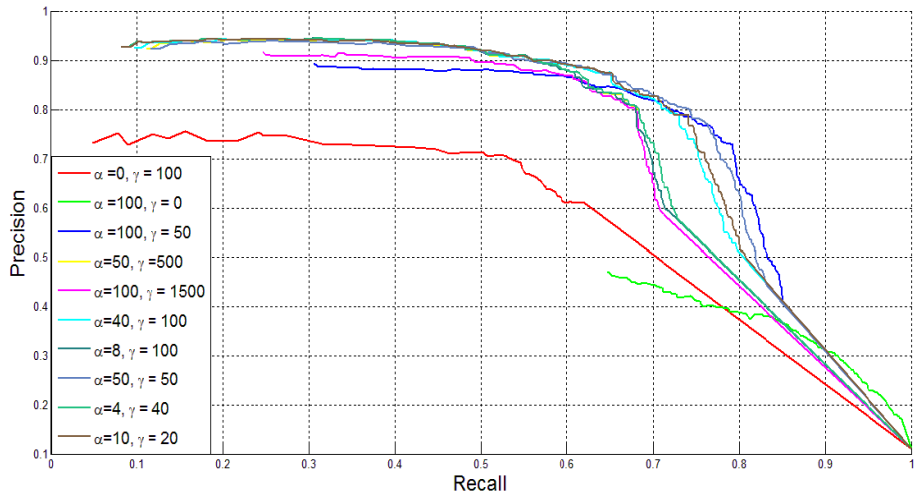

Fig. 8 The parameters tuning.

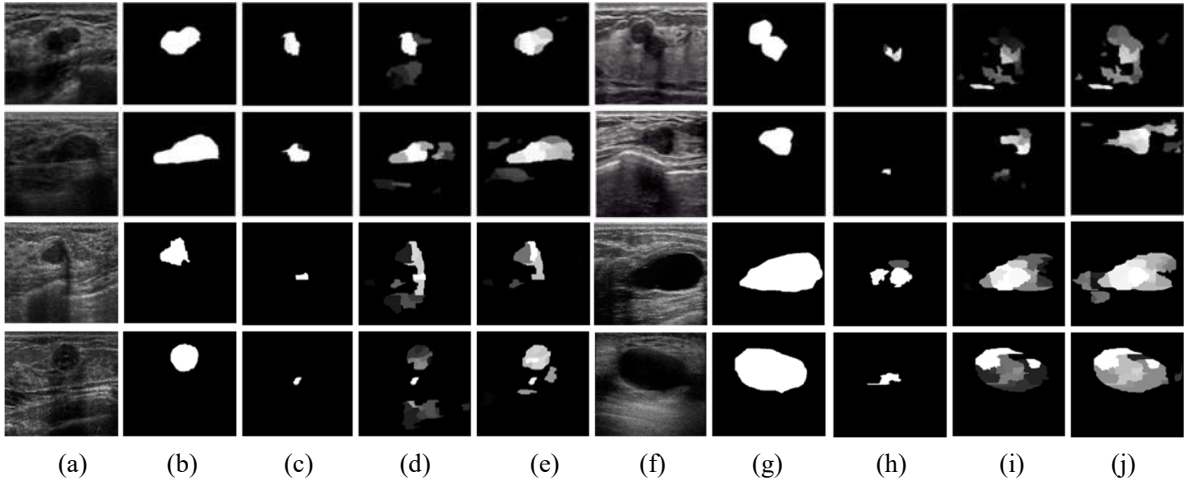

(a) (b) (c) (d) (e) (f) (g) (h) (i) (j)

Fig. 9. Effectiveness of the breast anatomy. (a) and (f) Original BUS images; (b) and (g) the ground truth; (c) and (h) without layers' information in both terms, OUR_NL; (d) and (i) without layers' information in BG term, OUR_NL_BG ; and (e) and (j) the saliency map with layers' information in both terms.

As shown in Fig. 8, the proposed approach achieves much better performance when the value of $\gamma$ is much bigger than that of $\alpha$; and when the value of $\alpha$ is less than 10, and $\gamma$ is less than 50, the performances are similar on the P-R curves; therefore, based on the minimum *MAE*, $\alpha = 4$ and $\gamma = 40$.

### 4.3. The effectiveness of the breast anatomy

Here, it compares the methods without the layers' information in the FG and BG generation. In our methods, $\alpha$ and $\gamma$ are set to 4 and 40 respectively. As shown in Figure 9, the proposed method without layers' information in both terms, abbreviated as OUR_NL, will fail to locate the tumor or most parts of the tumor will miss (see Figs. 9 (c) and (h)); and the proposed method without layers' information in the BG term, abbreviated as OUR_NL_BG, will miss some parts of the tumor (see the 1st and 2nd rows of Figs. 9 (d) and (i)) and cannot concentrate the high saliency values on the salient objects (see the 3rd and 4th rows of Figs. 9 (d) and (i)). The overall performances of OUR_NL, OUR_NL_BG and OURS in Figs. 13-14 demonstrate that the proposed method with the layers' information in the two terms is more robust than that without it.

### 4.4. The effectiveness of the new objective function

We illustrate the effectiveness of the new objective function by two category samples. 1) apply FG and BG generated in the method as the weighted map and NC map to the objective function in [34] which is one penalty objective function, abbreviated as OUR_OPT. Fig.10 shows some comparable samples results. The OUR_OPT method always locates the tumor position correctly and generates good saliency map on the images with large tumors (see Figs. 10 (a) and (c)). However, it will make the other non-tumor regions have high saliency values on the images with small tumors (see Figs. 10 (e) and (g)). The method with the proposed objective function can concentrate on high saliency values on the tumor regions and low values on the other regions. In addition, we apply the new method to the image without tumor and compare the result with that of RRWR[25], SMTD[7], HFTSE[34] and OUR_OPT. The sample results are shown in Fig. 11. The saliency maps are normalized to [0-1]. Fig.11 shows that the two penalty terms optimization framework can generate much more accurate saliency map than that of others.

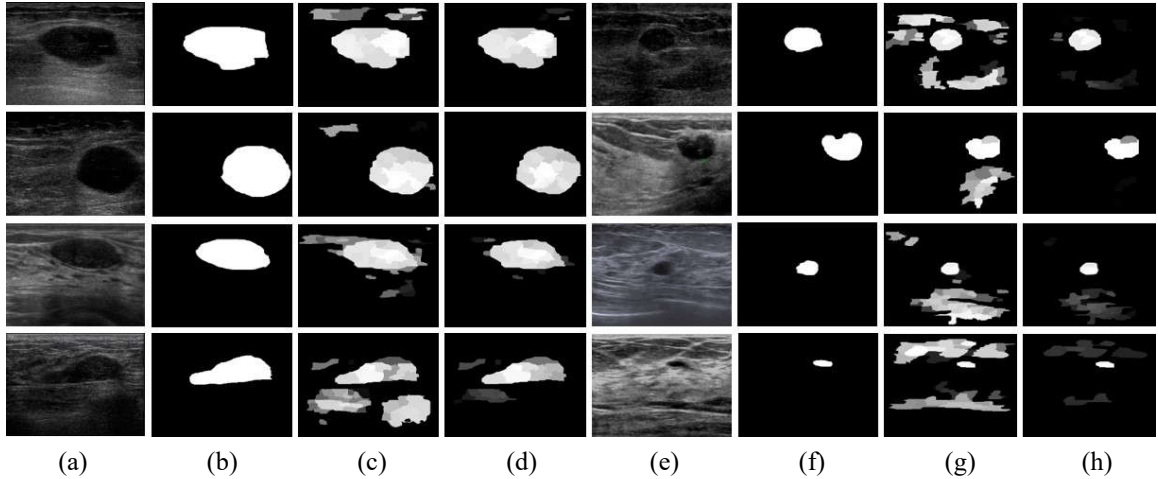

Fig. 10. Effectiveness of the new objective function. (a) and (e) original images; (b) and (f) the ground truth; (c) and (g) saliency maps obtained by OUR_OPT; and (d) and (h) saliency maps generated by the proposed method.

### 4.5. Overall performance

The proposed method is compared with most recently published methods SMTD [7], OMRC [27], MR [29], RRWR [25], HFTSE [34] and three models generated by the proposed method with different components in the optimization framework. RRWR, MR and OMRC are the bottom-up models and achieve good performances on the natural images. SMTD is the directly mapping method for tumor saliency estimation, and HFTSE is an optimization model to determine the existence of tumor and estimates tumor saliency for the image having tumors. OUR_NL is the two-penalty objective function with FG and BG maps in HFTSE; and OUR_NL_BG is the two-penalty objective function with the layered FG and the BG maps in HFTSE; and OUR_OPT is a one-penalty objective function with the layered FG and BG in HFTSE.

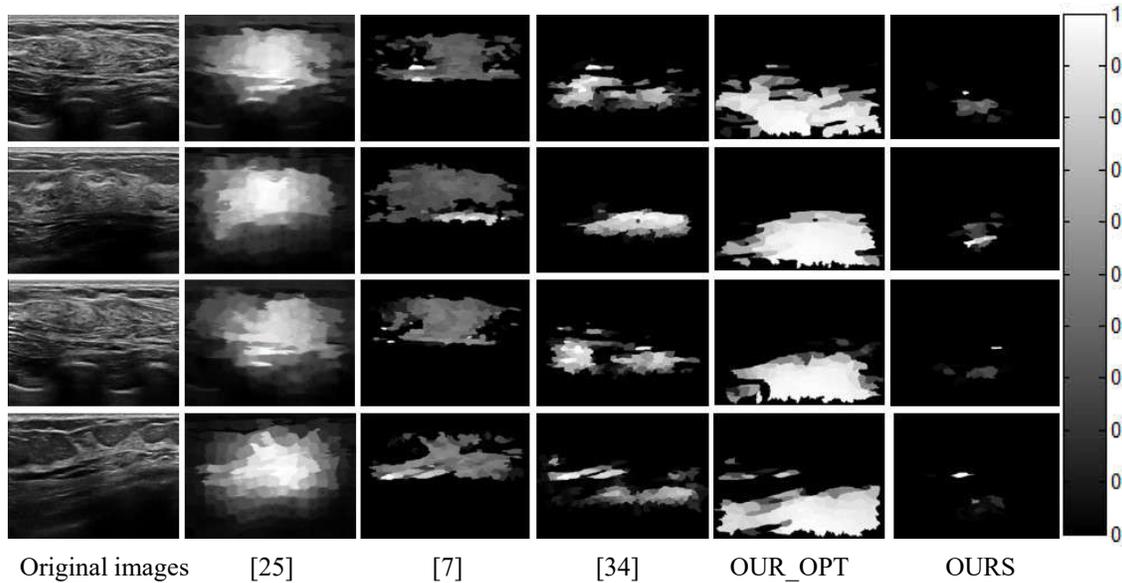

Fig. 11. Effectiveness of the new objective function on images without tumors. The region with higher intensity indicates the region with higher possibility belongs to a tumor.

Fig. 12 shows the comparison results of nine models using the same samples. The proposed method and other two models OUR_NL_BG, OUR_OPT can locate the tumors accurately; especially, for the image with the big or small tumors. OUR_NL_BG model can generate a similar saliency map as the proposed method; however, it would miss some part of the tumor as described in section 3.3. OUR_OPT model can highlight the non-tumor regions as well as the tumor regions. Therefore, this model can achieve higher recall ratio, as shown in Fig. 14. OUR_NL model will force the regions with a very high value in FG and very low value in BG to have high saliency value. Thus, it will make a very small part of the object to be the salient object. It will cause a higher precision ratio but lower recall ratio as shown in Fig. 14. HFTSE would miss parts of large tumors and miss the entire object in the image with low contrasts (see the 5$^{th}$ row of Fig. 12). SMTD would miss the object in the images with very big or very small tumors (see the 5$^{th}$, 8$^{th}$ and 10$^{th}$ rows of Fig. 12 and make the surround dark regions have high saliency values. OMTC, MR and RRWR worked better on the image with large tumors than that with small ones, even missed the small tumors (see the 5$^{th}$ row in Fig. 12); moreover, these methods made the background regions around the tumors have higher saliency values. This situation will make these methods have higher recall ratios but lower precision ratios.

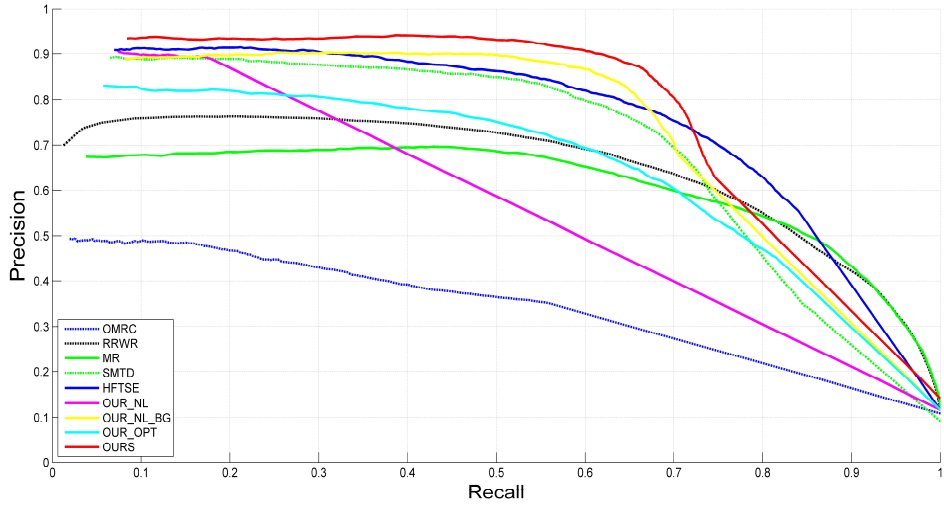

Fig 13. Precision-Recall curves of applying the nine models.

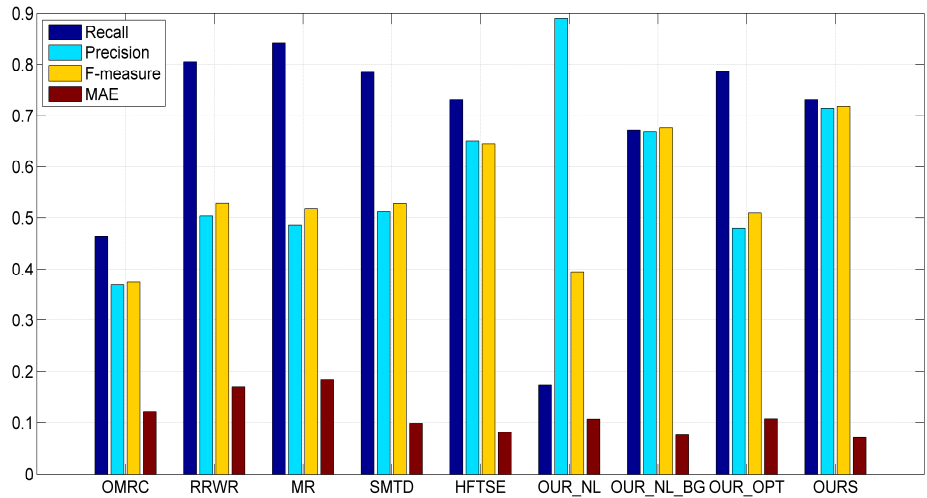

Fig. 14. The $F_{meansure}$, mean precision and recall ratios, and *MAE* of applying the nine models.

The overall performances of the nine models are shown in Figs. 13-14 using the metrics *MAE* values, $F_{meansure}$ values, and P-R curves. As shown in Fig. 13, the proposed method, noted as OURS, achieves a competitive P-R curve and the highest $F_{meansure}$ and the lowest *MAE*. As discussed, SMTD, MR, RRWR and HFTSE can obtain relatively high average recall ratios, but the precision ratios and F-measures are quite low. The reason is that these methods make the tumor and its surrounding background have high saliency values. OUR_NL only highlights a small part of the tumor as a salient object; and can achieve the highest precision ratio and the lowest recall ratio.

## 5. Conclusion

In this paper, we propose a novel optimization model to estimate tumor saliency by integrating the knowledge of breast anatomy for breast ultrasound images. There are four main contributions in this work: (1) breast anatomy is modeled and integrated into the proposed TSE framework; (2) more accurate foreground and background maps are generated by using the breast anatomy; especially for the images with large or small tumors; (3) the new objective function can handle the BUS images without breast tumors; and (4) the proposed method outperforms eight state-of-the-art TSE models on two datasets. In the future, we will focus on generalizing the proposed breast anatomy modeling approach to other image modalities and diseases.

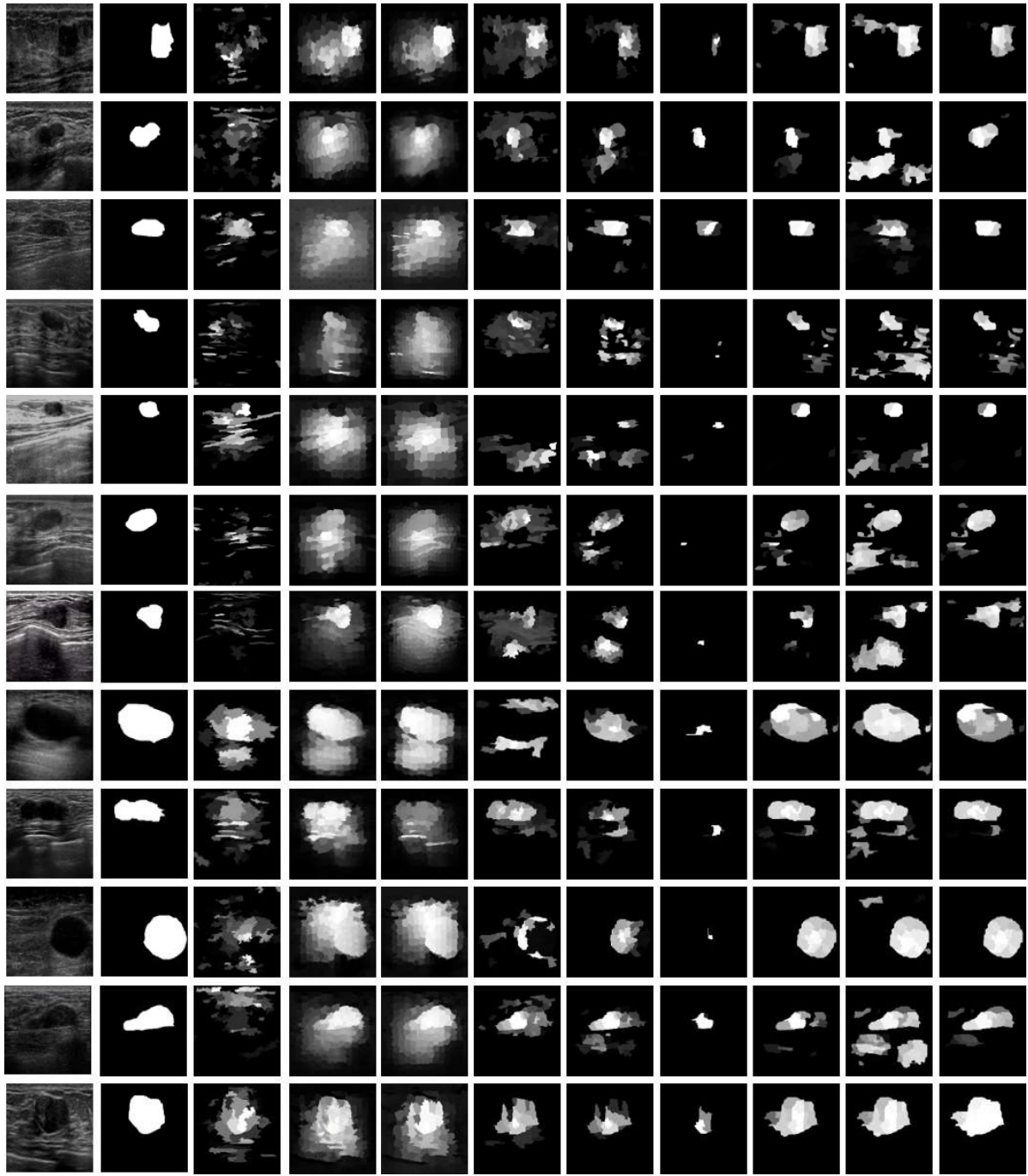

Fig. 12. Visual effects of detecting saliency maps of some example images by the nine methods.